\pdfoutput=1

\documentclass[11pt]{article}

\usepackage{naacl2021}

\usepackage{times}
\usepackage{latexsym}

\usepackage[T1]{fontenc}

\usepackage[utf8]{inputenc}

\usepackage{microtype}

\usepackage{listings}
\usepackage{graphicx}
\usepackage{url}

\PassOptionsToPackage{hyphens}{url}
\usepackage{hyperref}

\usepackage{booktabs}
\usepackage{multirow}
\usepackage{multicol}

\usepackage{array}
\newcolumntype{?}{!{\vrule width 1.5pt}}

\title{Does Dialog Length matter for Next Response Selection task? \\ An Empirical Study.}

\author{
  Jatin Ganhotra \\
  IBM Research \\
  {\tt jatinganhotra@us.ibm.com}
  \\\And
  Sachindra Joshi\\
  IBM Research\\
  {\tt jsachind@in.ibm.com}
  }
  
\begin{document}
\maketitle
\begin{abstract}

In the last few years, the release of BERT, a multilingual transformer based model, has taken the NLP community by storm. BERT-based models have achieved state-of-the-art results on various NLP tasks, including dialog tasks. One of the limitation of BERT is the lack of ability to handle long text sequence. By default, BERT has a maximum wordpiece token sequence length of 512. Recently, there has been renewed interest to tackle the BERT limitation to handle long text sequences with the addition of new self-attention based architectures. However, there has been little to no research on the impact of this limitation with respect to dialog tasks. Dialog tasks are inherently different from other NLP tasks due to: a) the presence of multiple utterances from multiple speakers, which may be interlinked to each other across different turns and b) longer length of dialogs. In this work, we empirically evaluate the impact of dialog length on the performance of BERT model for the Next Response Selection dialog task on four publicly available and one internal multi-turn dialog datasets. We observe that there is little impact on performance with long dialogs and even the simplest approach of truncating input works really well.

\end{abstract}

\section{Introduction}
BERT is a bidirectional model based on the transformer architecture \cite{devlin-etal-2019-bert} and is pre-trained on two unsupervised tasks: masked language modeling (MLM) and next sentence prediction (NSP). The pre-trained BERT model can be fine-tuned later on different downstream specific tasks such as sentiment classification, question answering, natural language inference etc. Recently, BERT-based models have been used for dialog tasks where they have shown great promise \cite{whang2020effective, gu2020speaker}.

However, there are still open questions and concerns with respect to using these models for dialog tasks. Various changes have been proposed for adapting BERT-based models to dialog tasks, e.g. \citet{gu2020speaker} propose Speaker-Aware BERT (SA-BERT) model to incorporate the notion of different speakers associated with various utterances in the dialog. The SA-BERT model performs substantially better than the default BERT model. \citet{sankar-etal-2019-neural} explore how the transformer-based seq2seq models use the available dialog history by studying its sensitivity to artificially introduced unnatural changes or perturbations to their context at test time. They observe that the transformer-based models are rarely sensitive to most perturbations such as missing or reordering utterances, shuffling words, etc.

One of the key questions, which has not received much attention from the community and left unnoticed is: impact of \textbf{dialog length} on performance of BERT-based models. In a dialog (especially goal-oriented dialog), various utterances are interlinked with each other and a common belief state is shared and updated as the dialog progresses. Traditionally, BERT-based models handle the input as a long concatenated sequence and insert special symbols like [SEP] tokens to mark the beginning and end of each individual utterance. However, \citet{ganhotra-etal-2020-conversational} show that real-world dialogs can be much longer than the BERT input max\_seq\_length of 512 tokens, and hence the entire dialog context can not be fed directly to the BERT model. 

By default, BERT has a maximum wordpiece token sequence length of 512. To address the input limitation, many different approaches there have been proposed. These approaches fall broadly in two categories: a) keep the BERT architecture same and b) replace the self-attention mechanism with new attention-based approaches. We discuss these approaches in more detail in Section \ref{bert-limit-approaches}. Usually, the key information of an article is at the beginning and end for classification (text/ news/ document) tasks. The same is true for dialog tasks as well, especially goal-oriented dialog. If we look at the anatomy of a dialog, the initial utterances set the overall scope of the conversation i.e. topic being discussed, or concern being raised in technical support, or primary intent of the request in a customer care setting. The next set of utterances involve back-and-forth between the speakers for finalizing the core specific details and the last few utterances conclude the conversation usually with a summary and/or confirmation of the task completion.

However for a long conversation, it's hard to conclusively say that the key information is at both the beginning and the end. It's possible that as the conversation progressed, the topic of the conversation may have shifted away from the original topic. In this work, we take an empirical approach to understand the impact of dialog length on BERT performance for the next response selection task. We experiment with four publicly available and one internal multi-turn dialog datasets, and share our findings in Section \ref{experiments}.

\begingroup
\setlength{\tabcolsep}{4pt} 

\begin{table*}[t]
  \centering
  \begin{tabular}{l|c|ccc|c|c}
  \toprule
  \multicolumn{2}{c|}{Dataset}                      & Train & Valid & Test & \# pos & > limit \\
  \hline
  \multirow{2}*{Ubuntu V1}  & pairs  & 1M    & 0.5M  & 0.5M & \multirow{2}*{49.9k} & \multirow{2}*{2218(4.4\%)} \\
                            & positive:negative  & 1: 1  & 1: 9  & 1: 9 & & \\
  \hline
  \multirow{2}*{Ubuntu V2}  & pairs  & 1M    & 195k  & 189k & \multirow{2}*{18.9k} & \multirow{2}*{882(4.7\%)} \\
                            & positive:negative  & 1: 1  & 1: 9  & 1: 9  & & \\
  \hline
  \multirow{2}*{Douban}     & pairs  & 1M    & 50k   & 10k  & \multirow{2}*{667} &  \multirow{2}*{36(5.4\%)}\\
                             & positive:negative & 1: 1  & 1: 1  & 1.2: 8.8  & & \\
  \hline
  \multirow{2}*{E-commerce}  & pairs & 1M    & 10k   & 10k  & \multirow{2}*{969} &  \multirow{2}*{2(0.2\%)}\\
                            & positive:negative  & 1: 1  & 1: 1  & 1: 9  & & \\
  \hline
  \hline
  \multirow{2}*{Tech-Support*} & pairs  &  208k   & 136k  & 135k  & \multirow{2}*{13.5k} &  \multirow{2}*{5521(40.8\%)}\\
                            & positive:negative  & 1: 1  & 1: 1  & 1: 9  & & \\

  \bottomrule
  \end{tabular}
  \caption{Statistics of the datasets. For each dataset, including the internal Tech-Support dataset, we provide details on the total number of train, valid and test pairs and the positive:negative sample ratio. The column '\# pos' refers to the total number of positive samples in the test set. The column '\textit{> limit}' refers to the count and (\%age) of samples where the input length is > 512.}
  \label{datasets-table}
\end{table*}
\endgroup


\section{Dealing with input > max\_seq\_length}
\label{bert-limit-approaches}
BERT has a maximum wordpiece token sequence length of 512. The magic number 512 was selected to reach a balance between performance and memory usage, as the self-attention mechanism scales quadratically in the sequence length. Thus, for longer input sequences with more than 512 word-piece tokens, which happens often in real-world dialog datasets (Section \ref{datasets}), the entire dialog might not fit within the limit. Several approaches have been proposed to circumvent/ tackle the max\_seq\_length ($\mathcal{T}$) limit of 512. These approaches can be categorized in two groups: a) keep BERT architecture same, and b) replace self-attention with a different attention-based approach. We discuss both of these approaches below:

\subsection{Group \#1: Keep BERT architecture}
There are four approaches in this group which try to circumvent the BERT max\_seq\_length limit: truncation, hierarchical methods, strides and combining predictions from input subsets. 

\subsubsection{Truncation methods}
\label{trunc-strategy}
There are different methods to truncate input text sequence to fit the BERT max\_seq\_length.
\begin{enumerate}
    \itemsep0em 
    \item head-only: Keep the first 512 tokens
    \item tail-only: Keep the last 512 tokens
    \item hybrid: Select first $n$ and last $512 - n$ tokens
    \item longest-first\footnote{\url{https://huggingface.co/transformers/preprocessing.html}}: Truncate token by token, removing a token from the longest sequence in the pair of input sequences
    \item speaker-aware disentanglement: Use a heuristic to select a subset of utterances from dialogs with more than 2 speakers \cite{gu2020speaker} e.g. DSTC 8-Track 2 \cite{kim2019eighth}  
\end{enumerate}

For hybrid and longest-first approaches, \textit{n} is selected based on the task and dataset, e.g. \citet{sun2019fine} select the first 128 and the last 382 tokens. 

\subsubsection{Hierarchical methods}
The input text (length \textit{L}) is divided into \textit{k = L / 512} fractions, which are fed into BERT to obtain the representations for the \textit{k} fractions. The hidden state of the [CLS] token is considered as the representation for the fraction. The final representation for the input is achieved by using mean pooling, max-pooling, or self-attention on the representations of all the fractions. However, this may not give a good representation of the entire input text. And the more the number of fractions, the more diluted the representation will be for the entire input. 
\citet{sun2019fine} observe that the hybrid truncation performs better than hierarchical approaches and achieves the best performance.





\subsubsection{Combine predictions from input subsets}
In this approach, the long input is simply split into sub-inputs to get a prediction for each sub-input. The predictions for all sub-inputs are then combined to get an overall prediction. \citet{yang-etal-2019-end} use this approach for open-ended question answering based on Wikipedia articles. They segment documents into paragraphs or sentences and then score only these smaller pieces. The final softmax layer over different answer spans is removed to allow comparison and aggregation of results from different segments.


\subsection{Group \#2: New architectures}
Recently, many new architectures have been proposed to address the challenge of scaling input length in the standard Transformer. 
Transformer-XL \cite{dai2019transformer} introduce segment-level recurrence mechanism and relative positional encoding scheme. 
They reuse the hidden states for previous segments as memory for the current segment, to build a recurrent connection between the segments.


Longformer \cite{beltagy2020longformer} replaces the standard quadratic self-attention with an attention mechanism that scales linearly with sequence length. They propose a combination of a windowed local-context self-attention and an end-task motivated global attention to encode inductive bias about the task.
Reformer \cite{Kitaev2020Reformer:} propose locality-sensitive hashing to reduce the sequence-length complexity and approximate the costly softmax computation in the full dot-product attention computation. Similar to Reformer, Performer \citet{choromanski2020rethinking} estimate the softmax attention by using a Fast Attention Via positive Orthogonal Random features approach. 

Big Bird \cite{zaheer2020big} propose a sparse attention mechanism that reduces the quadratic self-attention dependency to linear. \citet{ainslie-etal-2020-etc} introduce global-local attention mechanism between global tokens and regular input tokens. They combine the idea of Longformers and Randomized Attention which reduces quadratic dependency on the sequence length to linear, but requires additional layers for training.

\subsection{Current status w.r.t Dialog tasks}
Today, from the two groups mentioned above, these approaches are not used often for various NLP tasks as the number of articles longer than max\_seq\_length are very few (Table 1 \cite{sun2019fine}). Further, none of the new architectures from Group \#2 have been used for any dialog task so far. Even from Group \#1, only the truncation method has been used for next response selection task as BERT truncates input by default if it's longer than the max\_sequence\_length. The recent state-of-the-art model SA-BERT \cite{gu2020speaker} on multiple next response selection datasets uses truncation for handling longer dialogs. Hence, we use the truncation strategy \textit{'tail-only'} for our experiments to evaluate the impact of dialog length on the performance. We choose the \textit{'tail-only'} strategy because as the dialog progresses over multiple turns, it's likely that the recent utterances capture the most important information required for the end task of recommending next response.

\begin{table*}[ht]
  \small
  \centering
  \begin{tabular}{c?c|c|c|c|c?c|c|c|c|c}
  \toprule
     \multicolumn{6}{c|}{\textit{Ubuntu Corpus V1}} & \multicolumn{5}{c}{\textit{Ubuntu Corpus V2}} \\
    \hline
     \textit{len} & \# (\%) & $R_2@1$ & $R_{10}@1 $ & $R_{10}@2 $ & $R_{10}@5 $ & \#(\%) & $R_2@1$ & $R_{10}@1 $ & $R_{10}@2 $ & $ R_{10}@5 $\\
  \hline
\textit{all} & 49998 (100\%) & 0.951 & 0.813 & 0.9   & 0.976 & 18920 (100\%)  &           0.955 & 0.798 & 0.902 & 0.982 \\

\textit{< 512} & 47780 (95.6\%) & 0.951 & 0.808 & 0.897 & 0.976 & 18038 (95.3\%) & 0.953 & 0.796 & 0.901 & 0.982 \\

\textit{> 512} & 2218 (4.4\%) & 0.976 & 0.9 & 0.956 & 0.989  & 882 (4.7\%) & 0.967 & 0.837 & 0.923 & 0.991 \\

  \bottomrule
  \end{tabular}

\hfill \break

  \begin{tabular}{c?c|c|c|c|c|c?c|c|c|c|c}
  \toprule
     \multicolumn{7}{c|}{\textit{Douban Corpus}} & \multicolumn{5}{c}{\textit{Ecommerce Corpus}} \\
    \hline
     \textit{len} & \# (\%) & MAP &  MRR & $R_{10}@1 $ & $R_{10}@2 $ & $R_{10}@5 $ & \#(\%) &  MRR & $R_{10}@1 $ & $R_{10}@2 $ & $ R_{10}@5 $\\
  \hline
\textit{all} & 667 (100\%) & 0.438 & 0.477 & 0.267 & 0.451 & 0.78 & 969 (100\%)    & 0.533 & 0.327 & 0.524 & 0.829 \\

\textit{< 512} & 633 (94.9\%) & 0.441 & 0.48  & 0.272 & 0.455 & 0.784 & 967 (99.8\%) & 0.533 & 0.327 & 0.524 & 0.828 \\ 

\textit{> 512} & 36 (5.4\%) & 0.391 & 0.444 & 0.222 & 0.417 & 0.722 & 2 (0.2\%) & 0.625 & 0.5 & 0.5 & 1 \\

  \bottomrule
  \end{tabular}

\hfill \break

\begin{tabular}{c?c|c|c|c|c}
\toprule
\multicolumn{6}{c}{\textit{Tech-support}} \\
\hline
\textit{len} & \# (\%) & MRR & $	R_{10}@1 $ & $	R_{10}@2 $ & $ R_{10}@5$ \\
\hline
all & 13536 (100\%) & 0.93  & 0.885 & 0.947 & 0.992 \\
< 512 & 8015 (59.2\%)   & 0.929 & 0.883 & 0.948 & 0.993 \\
> 512 & 5521 (40.8\%)   & 0.931 & 0.888 & 0.945 & 0.991 \\
\hline
\end{tabular}
\caption{Evaluation results of BERT on all multi-turn next response selection datasets. The column \textit{len} refers to the dialog length. The column \# (\%) refers to the number of test dialogs per subset and their \%age in comparison to the overall test set. For rows with \textit{len > 512}, dialogs were truncated from beginning i.e. tail-only truncation.}
\label{all-results-table}
\end{table*}


\section{Experiments}
\label{experiments}
The next response selection task is defined as follows: Given a dialog dataset $\mathcal{D}$, an example is denoted as a triplet \textless $d,r,l$ \textgreater, where $c = \{u_1,u_2,...,u_n\}$ represents the dialog context with $n$ utterances($u$), $r$ is a response candidate, and $l \in \{0,1\}$ denotes a label. When $r$ is the correct response for $c$, then $y=1$; else $y=0$.

\subsection{Datasets}
\label{datasets}
We experiment with four public and one internal multi-turn response selection dialog datasets across different domains (system troubleshooting, social network, e-commerce and technical support): 
Ubuntu Dialogue Corpus V1 \cite{lowe-etal-2015-ubuntu}, Ubuntu Dialogue Corpus V2 \cite{lowe2017training}, Douban Conversation Corpus \cite{wu-etal-2017-sequential}, E-commerce Dialogue Corpus \cite{zhang2018modeling} and our internal Tech-Support dataset. Statistics of the datasets are provided in Table \ref{datasets-table}. The Tech-Support dataset was generated following \citet{lowe-etal-2015-ubuntu} from real-world conversations between users and human agents on technical support.

We observe that the \%age of long dialogs in publicly available benchmarks is very small (5\%), while its much higher (40\%) for our internal Tech-Support dataset. This shows that existing public benchmarks for the next response selection task are not a good representative for real-world dialogs, in the context of overall dialog length.

\subsection{Evaluation Metrics}
For evaluation, we use $R_n@k$ where the model is asked to select the $k$ most best-matched responses from $n$ available candidates, and it is correct if the correct response is among these $k$ responses. In addition to $R_n@k$, we also use Mean Reciprocal Rank (MRR) and Mean Average Precision (MAP). MRR captures the rank of the first relevant item, while MAP captures whether all of the relevant items are ranked highly, which is important for the Douban corpus, as there are multiple correct candidates for a dialog context in the test set.

\subsection{Experimental Results}
We use the uncased BERT-base model\footnote{\url{https://storage.googleapis.com/bert_models/2018_10_18/uncased_L-12_H-768_A-12.zip}} for our experiments. For fine-tuning, all hyper-parameters of the original model were followed \cite{devlin-etal-2019-bert}.
We divide the test set for each dataset in two groups based on dialog length: a) < 512 and b) > 512. We compute the evaluation metrics for both groups in addition to the overall test set. The results are provided in Table \ref{all-results-table}. For the 2nd group, where input > 512, we use the '\textit{tail-only}' truncation strategy explained in Section \ref{trunc-strategy}, because as the conversation progresses, the most recent utterances will be more important for the next response selection task.

We observe that the performance of BERT model on the 2nd group (> 512) is comparable to the overall test set performance across all the datasets, including our internal Tech-Support dataset which has 40\% test samples which are longer than the BERT max\_sequence\_len 512. 
This implies that a pretrained BERT model deployed in production for the next response selection would not suffer on performance if the input was longer than the BERT limit and the most obvious strategy (\textit{tail-only}) to use the most recent utterances from dialog will work really well.

\section{Conclusion}
\label{conclusion}
With the rising popularity of BERT-based models in the NLP community, we empirically evaluate the impact of dialog length on the performance of BERT in the context of the next response selection dialog task. We observe that existing public benchmarks for the next response selection task are not a good representative for real-world dialogs, in the context of overall dialog length. To our surprise, we notice that there is no performance drop for longer dialogs and even the simplest 'tail-only' truncation approach performs really well.

\newpage

\bibliography{anthology,custom}
\bibliographystyle{acl_natbib}

\end{document}